\documentclass[10pt,journal,compsoc]{IEEEtran}
\pdfoutput=1
\usepackage[utf8]{inputenc}
\usepackage{amsmath,graphicx}
\usepackage{epstopdf}
\usepackage{algorithm,algorithmic}
\usepackage{setspace}
\usepackage{cite}
\usepackage[colorlinks,linkcolor=blue,anchorcolor=blue,citecolor=blue]{hyperref}
\usepackage{subfig}
\usepackage{subfloat}
\usepackage{multirow,booktabs}
\usepackage{fontenc}

\usepackage{amsmath,amssymb,mathrsfs}
\usepackage[misc]{ifsym}
\usepackage{url}
\usepackage{bm}
\usepackage{bbding}
\usepackage{mathtools}
\usepackage{makecell}
\usepackage{amsfonts,amsthm}
\usepackage{hyperref}
\hypersetup{hypertex=true,
	colorlinks=true,
	linkcolor=red,
	anchorcolor=green,
	citecolor=blue}

\newtheorem{remark}{Remark}

\usepackage{array}
\usepackage{cases}
\newcolumntype{"}{@{\hskip\tabcolsep\vrule width 1pt\hskip\tabcolsep}}
\usepackage{pifont}
\usepackage{color}

\usepackage{amsmath,amsfonts,bm}

\def\eqref#1{equation~\ref{#1}}

\def\1{\bm{1}}

\def\vf{{\bm{f}}}

\def\vh{{\bm{h}}}

\def\vp{{\bm{p}}}

\def\vw{{\bm{w}}}
\def\vx{{\bm{x}}}

\def\mA{{\bm{A}}}

\DeclareMathAlphabet{\mathsfit}{\encodingdefault}{\sfdefault}{m}{sl}
\SetMathAlphabet{\mathsfit}{bold}{\encodingdefault}{\sfdefault}{bx}{n}

\newtheorem{proposition}{Proposition}
\newtheorem{theorem}{Theorem}

\begin{document}

\title{Neural Delay Differential Equations: System Reconstruction and Image Classification}

\author{Qunxi~Zhu,
 Yao~Guo,
 and~Wei~Lin,~\IEEEmembership{Senior Member,~IEEE}
\IEEEcompsocitemizethanks{\IEEEcompsocthanksitem 
Q. Zhu and Y. Guo are with the Research Institute of Intelligent Complex Systems, Fudan University, Shanghai 200433, China.\protect\\
E-mail: qxzhu16@fudan.edu.cn, yguo@fudan.edu.cn.
\IEEEcompsocthanksitem 
W. Lin is with the Research Institute of Intelligent Complex Systems, the MOE Frontiers Center for Brain Science and State Key Laboratory of Medical Neurobiology, the School of Mathematical Sciences, SCMS, SCAM, and CCSB, and the Shanghai Artificial Intelligence Laboratory, Fudan University, Shanghai 200433, China. \protect\\
E-mail: wlin@fudan.edu.cn. 
\IEEEcompsocthanksitem 
Corresponding authors: Q. Zhu, Y. Guo and W. Lin. \protect\\

}
\thanks{}}
\markboth{}
{Shell \MakeLowercase{\textit{et al.}}: Bare Demo of IEEEtran.cls for Computer Society Journals}

\IEEEtitleabstractindextext{

\begin{abstract}
Neural Ordinary Differential Equations (NODEs), a framework of continuous-depth neural networks, have been widely applied, showing exceptional efficacy in coping with representative datasets. Recently, an augmented framework has been developed to overcome some limitations that emerged in the application of the original framework. In this paper, we propose a new class of continuous-depth neural networks with delay, named Neural Delay Differential Equations (NDDEs). To compute the corresponding gradients, we use the adjoint sensitivity method to obtain the delayed dynamics of the adjoint. Differential equations with delays are typically seen as dynamical systems of infinite dimensions that possess more fruitful dynamics. Compared to NODEs, NDDEs have a stronger capacity for nonlinear representations. We use several illustrative examples to demonstrate this outstanding capacity. Firstly, we successfully model the delayed dynamics where the trajectories in the lower-dimensional phase space could be mutually intersected and even chaotic in a model-free or model-based manner. Traditional NODEs, without any argumentation, are not directly applicable to such modeling. Secondly, we achieve lower loss and higher accuracy not only for the data produced synthetically by complex models but also for the CIFAR10, a well-known image dataset. Our results on the NDDEs demonstrate that appropriately articulating the elements of dynamical systems into the network design is truly beneficial in promoting network performance.
\end{abstract}

}
\maketitle
\IEEEdisplaynontitleabstractindextext
\IEEEpeerreviewmaketitle

\section{Introduction}
\IEEEPARstart{I}n the past decade, machine learning techniques have achieved significant success in many scientific fields, with residual networks \cite{he2016deep} and their variants being particularly prominent. They can train very deep neural networks and have become the state-of-the-art architecture for a large number of tasks, such as image classification \cite{krizhevsky2012imagenet, he2016deep}, speech recognition \cite{xiong2018microsoft}, natural language processing \cite{vaswani2017attention, devlin2018bert}, playing Go \cite{silver2016mastering, silver2017mastering}, and protein structure prediction \cite{service2020game}. As a result of their widespread use, researchers have explored the underlying mechanisms that enable residual connections between layers to train very deep neural networks. 

Recent research has highlighted a strong correlation between neural networks and dynamical systems \cite{weinan2017proposal, li2017maximum, haber2017stable,chang2017multi, li2018optimal, lu2018beyond, weinan2019mean, chang2019antisymmetricrnn, ruthotto2019deep, zhang2019you, 2018Model, zhu2019detecting, tang2020introduction, chen2018neural}. This connection has led to the development of novel and efficient frameworks for neural networks, as well as the ability to solve ordinary and partial differential equations that were previously difficult to compute using traditional algorithms. A typical example is the Neural Ordinary Differential Equations (NODEs) framework, which treats the infinitesimal time of ordinary differential equations (ODEs) as the ``depth'' of a considered neural network~\cite{chen2018neural}. 

Though the NODEs have demonstrated clear advantages in modeling continuous-time datasets and continuous normalizing flows with constant memory cost \cite{chen2018neural}, their limitations in representing certain functions have also been studied \cite{dupont2019augmented}. For instance, NODEs cannot be directly applied to describe dynamical systems where the trajectories in the lower-dimensional phase space intersect. Additionally, they cannot model only a few variables from certain physical or physiological systems where time delay is a significant factor. These limitations can be attributed to the finite-dimensional characteristic of NODEs from a dynamical systems theory perspective.

In this article, we propose a novel framework of continuous-depth neural networks with delay, referred to as Neural Delay Differential Equations (NDDEs). We apply the adjoint sensitivity method to compute the corresponding gradients, where the obtained adjoint systems are also in a form of delay differential equations. The main virtues of the NDDEs include:
\begin{itemize}
 \item feasible and computable algorithms for computing the gradients of the loss function based on the adjoint systems,
 \item representation capability of the vector fields which allow the intersection of the trajectories in the lower-dimensional phase space, and
 \item accurate reconstruction of the complex dynamical systems with effects of time delays based on the observed time-series data. 
\end{itemize}

A previous version of this work was published at ICLR 2021 \cite{zhu2021neural}. This paper extends the conference version with several significant new contributions: a) We regard the time delay and the terminal time of NDDEs as the additional trainable parameters and derive the mathematical forms of their gradients for given loss functions; b) We provide new insights into the link between the NDDEs and NODEs that the NDDEs can be regarded as an infinitely-dimensional augmented NODEs; c) We analyze the computational complexity of the NDDE framework; d) We further conduct more extensive experiments, including the visualization of the decision boundaries for NODEs and NDDEs in different training epochs on a concentric dataset, the model-based or model-free system reconstructions without knowing the true time dealy, and the image classifications with fixed/trainable terminal time points; e) and we extend more details and intuitions on the experimental results as well as further discussions on the proposed framework.
\begin{figure*}[htb]
 \begin{center}
 \includegraphics[width=17cm]{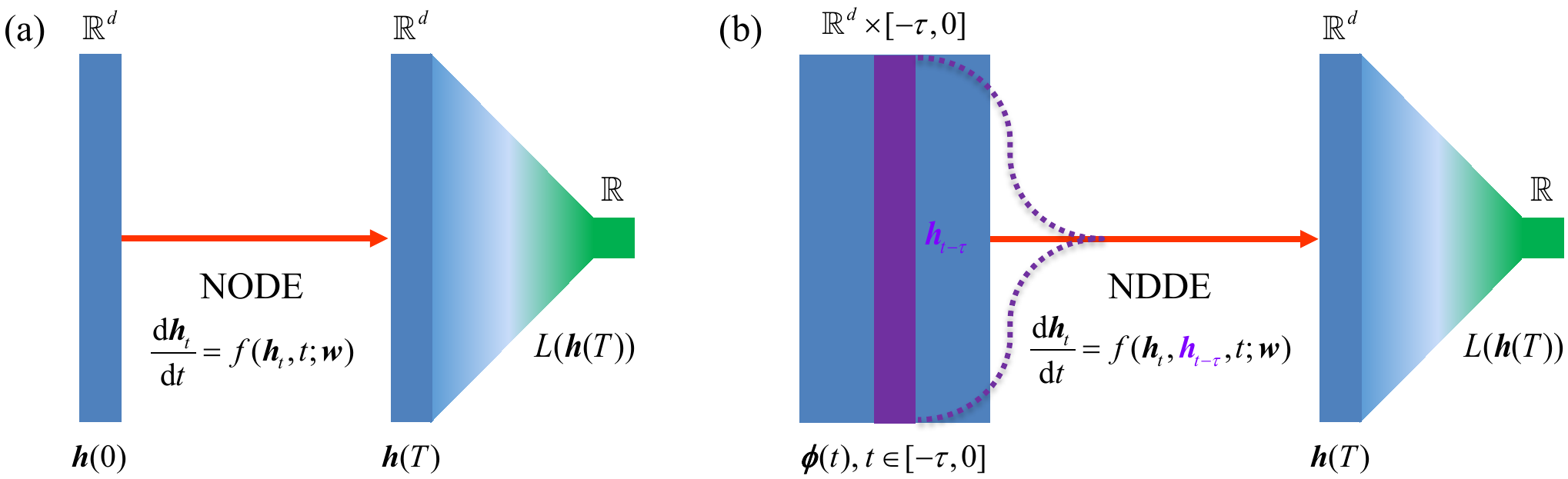}
 \end{center}
 \caption{Sketchy diagrams of the NODEs (a) and the NDDES (b), respectively, with the initial value $\vh(0)$ and the initial function $\bm{\phi}(t)$. The NODEs and the NDDEs act as the feature extractors, and the following layer processes the features with a predefined loss function.}
 \label{fig_diagram_ndde}
\end{figure*}

\section{Related Works}

{\bf NODEs} Inspired by the residual neural networks \cite{he2016deep} and the other analogous frameworks, the NODEs were established, which can be represented by multiple residual blocks as
\begin{equation*}
 \vh_{t+1} = \vh_t + \vf(\vh_t, \vw_t),
\end{equation*}
where $\vh_t$ is the hidden state of the $t$-th layer, $\vf(\vh_t; \vw_t)$ is a differential function preserving the dimension of $\vh_t$, and $\vw_t$ is the parameter pending for learning. The evolution of $\vh_t$ can be viewed as the special case of the following equation 
\begin{equation*}
 \vh_{t+\Delta t} = \vh_t + \Delta t \cdot \vf(\vh_t, \vw_t)
\end{equation*}
with $\Delta t=1$. As suggested in \cite{chen2018neural}, all the parameters $\vw_t$ are unified into $\vw$ for achieving parameter efficiency of the NODEs. This unified operation was also employed in the other neural networks, such as the recurrent neural networks (RNNs) \cite{rumelhart1986learning, elman1990finding} and the ALBERT \cite{lan2019albert}.
Letting $\Delta t \rightarrow 0$ and using the unified parameter $\vw$ instead of $\vw_t$, we obtain the continuous evolution of the hidden state $\vh_t$ as
\begin{equation*}
 \lim_{\Delta t \rightarrow 0} \dfrac{\vh_{t+\Delta t} - \vh_t}{\Delta t} = \dfrac{{\rm{d}}{\vh_t}}{{\rm{d}} t} = \vf(\vh_t, t; \vw),
\end{equation*}
which is in the form of ordinary differential equations. Actually, the NODEs can act as a feature extraction, mapping an input to a point in the feature space by computing the forward path of a NODE as:
\begin{equation*}
 \vh(T)= \vh(0) + \int_{0}^{T} \vf(\vh_t, t; \vw) d t, ~~\vh(0)=\mbox{input},
\end{equation*}
where $\vh(0)=\mbox{input}$ is the original data point or its transformation, and $T$ is the integration time (assuming that the system starts at $t=0$). 

 Under a predefined loss function $L(\vh(T))$, \cite{chen2018neural} employed the adjoint sensitivity method to compute the memory-efficient gradients of the parameters along with the ODE solvers. More precisely, they defined the adjoint variable, $\bm{\lambda}(t) = \dfrac{\partial L(\vh(T))}{\partial \vh(t)}$,

 whose dynamics is another ODE, i.e.,
 \begin{equation}
 \dfrac{{\rm{d}} \bm{\lambda}(t)}{{\rm{d}} t} 
 = -\bm{\lambda}(t)^{\top} \dfrac{\partial f(\vh_t, t; \vw)}{\partial \vh_t}.
 \end{equation}
 The gradients are computed by an integral as:
 \begin{equation}
 \dfrac{{\rm{d}} L}{{\rm{d}} \vw} = \int_{T}^0 -\bm{\lambda}(t)^{\top} \dfrac{\partial f(\vh_t, t; \vw)}{\partial \vw} {\rm{d}} t.
 \end{equation}
 
 Ref. \cite{chen2018neural} calculated the gradients by calling an ODE solver with extended ODEs (i.e., concatenating the original state, the adjoint, and the other partial derivatives for the parameters at each time point into a single vector). Notably, for the regression task of the time series, the loss function probably depends on the state at multiple observational times, such as the form of $L(h(t_0), h(t_1),...,h(t_n))$. Under such a case, we must update the adjoint state instantly by adding the partial derivative of the loss at each observational time point.

As emphasized in \cite{dupont2019augmented}, the flow of the NODEs cannot represent some functions omnipresently emergent in applications. Typical examples include the following two-valued function with one argument: $g_{\mbox{\tiny\rm 1-D}}(1) = -1$ and $g_{\mbox{\tiny\rm 1-D}}(-1) = 1$. Our framework desires to conquer the representation limitation observed in applying the NODEs. 

{\bf Optimal control} As mentioned above, a closed connection between deep neural networks and dynamical systems has been emphasized in the literature and, correspondingly, theories, methods, and tools of dynamical systems have been employed, e.g. the theory of optimal control \cite{weinan2017proposal, li2017maximum, haber2017stable, chang2017multi, li2018optimal, weinan2019mean, chang2019antisymmetricrnn, ruthotto2019deep, zhang2019you}. Generally, we model a typical task using a deep neural network and then train the network parameters such that the given loss function can be reduced by some learning algorithm. In fact, training a network can be seen as solving an optimal control problem on difference or differential equations ~\cite{weinan2019mean}. The parameters act as a controller with the goal of finding an optimal control to minimize/maximize some objective function. Clearly, the framework of the NODEs can be formulated as a typical problem of optimal control of ODEs. Additionally, the framework of NODEs has been generalized to the other dynamical systems, such as the Partial Differential Equations (PDEs) \cite{han2018solving, long2018pde, long2019pde, ruthotto2019deep, sun2020neupde} and the Stochastic Differential Equations (SDEs) \cite{lu2018beyond, sun2018stochastic, liu2019neural}, where the theory of optimal control has been completely established. It is worthwhile to mention that the optimal control theory is tightly connected with and benefits from the method of the classical calculus of variations \cite{liberzon2011calculus}. We also will transform our framework into an optimal control problem, and finally solve it using the method of the calculus of variations.

\section{The framework of NDDEs}

\subsection{Formulation of NDDEs}

\begin{figure}[htb]
\centering

	\includegraphics[width=0.42\textwidth]{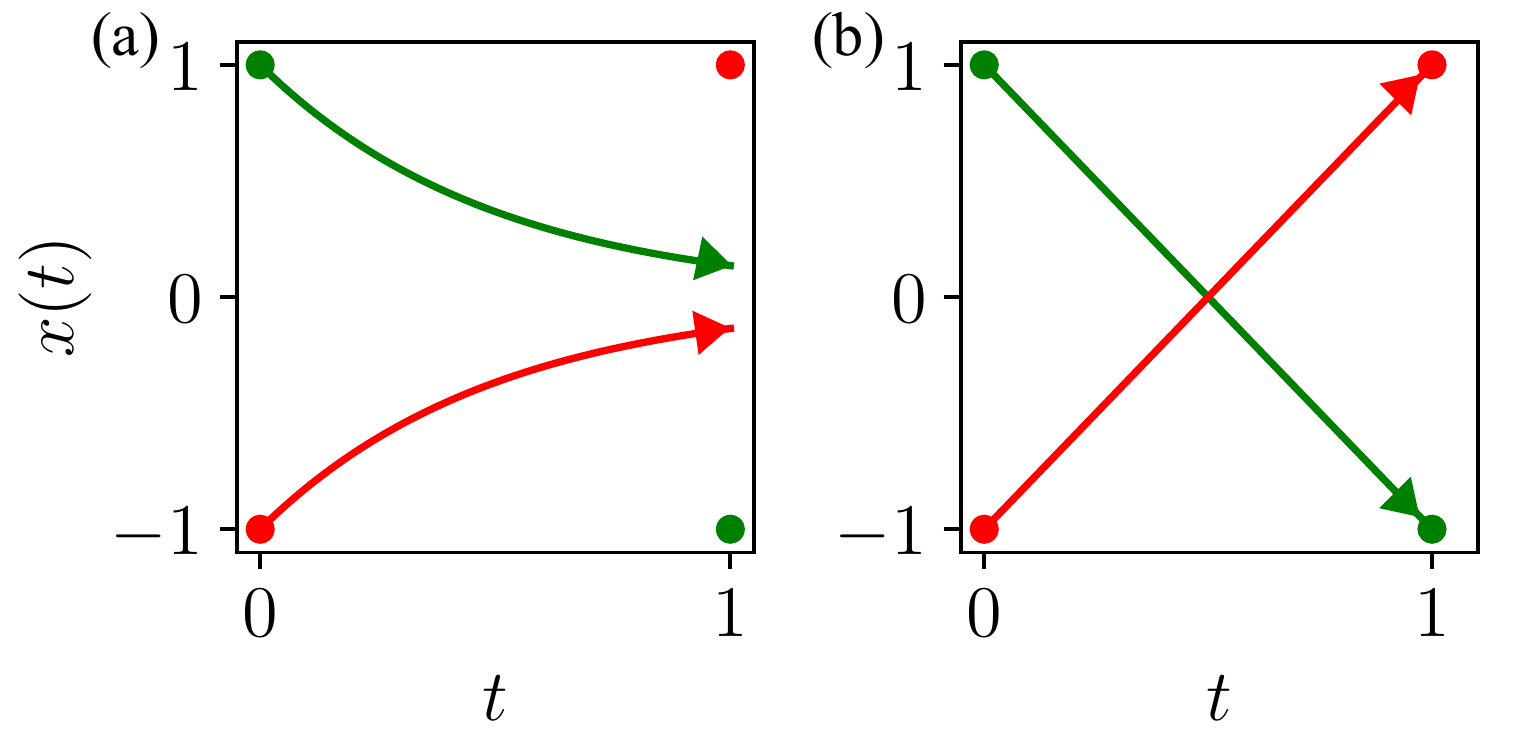}
	\caption{(Right) Two continuous trajectories generated by the DDEs are intersected, mapping -1 (resp., 1) to 1 (resp., -1), while (Left) the ODEs cannot represent such mapping.}
 \label{fig_01}
\end{figure}

In this section, we establish a framework of continuous-depth neural networks. To this end, we first introduce the concept of delay differential equations (DDEs). The DDEs are always written in a form where the derivative of a given variable at time $t$ is affected not only by the current state of this variable but also the states at some previous time instants or time durations \cite{erneux2009applied}. Such kind of delayed dynamics plays an important role in the description of the complex phenomena emergent in many real-world systems, such as physical, chemical, ecological, and physiological systems. In this article, we consider a system of DDE with a single time delay:
\begin{equation}\label{eq:ndde}
 \left\{
 \begin{array}{ll}
 \dfrac{{\rm{d}}{{\vh}_t}}{{\rm{d}} t} = f(\vh_t, \vh_{t-\tau}, t; \vw), & t>=0, \\
 \vh(t)={ {\bm{\phi}} }(t), & t<=0,
 \end{array}
 \right.
\end{equation}
where the positive constant $\tau$ is the time delay and $\vh(t)={\bm{\phi}}(t)$ is the \textit{initial function} before the time $t=0$. Clearly, in the initial problem of ODEs, we only need to initialize the state of the variable at $t=0$ while we initialize the DDEs using a continuous function. Here, to highlight the difference between ODEs and DDEs, we provide a simple example:
\begin{equation*}
 \left\{
 \begin{array}{ll}
 \dfrac{{\rm{d}}{x_t}}{{\rm{d}} t} = - 2 x_{t-\tau}, & t>=0, \\
 x(t)=x_0, & t<=0.
 \end{array}
 \right.
\end{equation*}
where $\tau=1$ (with time delay) or $\tau=0$ (without time delay). As shown in Figure $\ref{fig_01}$, the DDE flow can map -1 to 1 and 1 to -1; nevertheless, {this cannot be made for the ODE whose trajectories are not intersected with each other in the $t$-$x$ space in Figure~\ref{fig_01}.
}

\subsection{Adjoint Method for NDDEs}

Assume that the forward pass of DDEs is complete. Then, we need to compute the gradients in a reverse-mode differentiation by using the adjoint sensitivity method \cite{chen2018neural, pontryagin1962mathematical}. We consider an augmented variable, named as \textit{adjoint} and defined as
\begin{equation}
 \bm{\lambda}(t) = \dfrac{\partial L(\vh(T))}{\partial \vh(t)},
\end{equation}
where $L(\cdot)$ is the loss function pending for optimization.
Notably, the resulting system for the adjoint is in a form of DDE as well. 

\begin{theorem} 
 \label{thm_01}
 (Adjoint method for NDDEs). Consider the loss function $L(\cdot)$. Then, the dynamics of adjoint can be written as
 \begin{equation}
 \label{adjoint}
	\small
 \left\{
 \begin{aligned}
 \dfrac{{\rm{d}} \bm{\lambda}(t)}{{\rm{d}} t} 
 = &-\bm{\lambda}(t)^{\top} \dfrac{\partial f(\vh_t, \vh_{t-\tau}, t; \vw)}{\partial \vh_t} \\
&- \bm{\lambda}(t+\tau)^{\top} \dfrac{\partial f(\vh_{t+\tau}, \vh_{t}, t; \vw)}{\partial \vh_t} \chi_{[0, T-\tau]}(t), ~t<=T\\
 \bm{\lambda}(T) = & \dfrac{\partial L(\vh(T))}{\partial \vh(T)},
 \end{aligned}
 \right.
 \end{equation}
where $\chi_{[0, T-\tau]}(\cdot)$ is a typical characteristic function. 
\end{theorem}

We provide two ways to prove Theorem \ref{thm_01}, which are, respectively, shown in Appendix. Clearly, when the delay $\tau$ approaches zero, the adjoint dynamics degenerate as the conventional case of the NODEs \cite{chen2018neural}. In addition, based on the adjoint Eq. $(\ref{adjoint})$, we have the following result, whose proof can be found in the Appendix.
\begin{proposition}
 \label{prop_grad}
	According to the adjoint Eq. $(\ref{adjoint})$, we have
 \begin{equation}
 \label{parasgrad}
\left\{
	\begin{aligned}
 		\dfrac{{\rm{d}} L}{{\rm{d}} \vw} &= \int_{T}^0 -\bm{\lambda}(t)^{\top} \dfrac{\partial f(\vh_t, \vh_{t-\tau}, t; \vw)}{\partial \vw} \rm{d} t,\\
		\dfrac{{\rm{d}} L}{{\rm{d}} \tau} &= \int_{T}^{0}\bm{\lambda}(t)^{\top} \dfrac{\partial f(\vh_t, \vh_{t-\tau}, t; \vw)}{\partial \vh_{t-\tau}} \dot{\vh}(t-\tau) \rm{d} t,\\
\dfrac{{\rm{d}} L}{{\rm{d}} \vh_0} & = \bm{\lambda}(0),\\
\dfrac{{\rm{d}} L}{{\rm{d}} T} & = \bm{\lambda}(T)^{\top}f(\vh_t, \vh_{t-\tau}, t; \vw).
	\end{aligned}
\right.
	\end{equation}
\end{proposition}
\begin{remark}
From Proposition \ref{prop_grad}, one can, according to the specific task, not only optimize the parameters in neural networks but also the time delay, the terminal time, and the initial state. If the initial function is a constant function, then $\dot{\vh}(t)\equiv 0$ for $t \in [-\tau, 0]$. In our experiments, the initial function can be modeled by a parameterized NODE, i.e., $\dot{\vh}(t) = \Phi(\vh, \vw_{\Phi})$ for $t \in [-\tau, 0]$ with $\vw_{\Phi}$ being the trainable parameter vector.
\end{remark}
\begin{remark}
Notably, the NDDEs $(\ref{eq:ndde})$ can be rewritten as a piece-wise augmented NODEs in : 
\begin{equation}
\label{eq:ndde_aug}
\left\{
 \begin{array}{ll}
 \dfrac{{\rm{d}}{{\vh}_t^{(n)}}}{{\rm{d}} t} = f\left(\vh_t^{(n)}, \vh_t^{(n-1)}, t; \vw\right), & t \in [(n-1)\tau, n\tau], \\
	\dfrac{{\rm{d}}{{\vh}_t^{(n-1)}}}{{\rm{d}} t} = f\left(\vh_t^{(n-1)}, \vh_t^{(n-2)}, t; \vw\right), & t \in [(n-1)\tau, n\tau], \\
~~~~\vdots&\\
\dfrac{{\rm{d}}{{\vh}_t^{(1)}}}{{\rm{d}} t} = f\left(\vh_t^{(1)}, \vh_t^{(0)}, t; \vw\right), & t \in [(n-1)\tau, n\tau], \\

{{\vh}_{t}^{(k)}} := \vh\left(t - (n-k)\tau\right), &k=1, 2..., n,\\
 \vh_t^{(0)}:={ {\bm{\phi}} }(t-n\tau), & k=0.
 \end{array}
 \right.
\end{equation} 
As the time $t$ approaches infinity, the integer $n$ approaches infinity as well, implying that the NDDEs $(\ref{eq:ndde})$ can be regarded as an infinitely-dimensional augmented NODEs. In our experiments, to reduce the computational cost suffering from the increasing high dimensionality of Eq. $(\ref{eq:ndde_aug})$, we typically use a small integer $n$ to learn the NDDE model and utilize the natural cubic spline (used in the work \cite{zhu2023leveraging} as well) to approximate the initial function, i.e., $\phi(t)\approx \textbf{Spline}(\hat{\vh}, t), t\in[-\tau, 0]$ with $\hat{\vh} =\left\{[t_0, \vh(t_0)[, [t_1, \vh(t_1)], ..., [t_m, \vh(t_m)]\right\}$ being the observable initial time series, satisfying $t_0\leq-\tau$ and $t_m\geq 0$.
\end{remark}
\begin{algorithm*}{htb}
 \caption{Piece-wise reverse-mode derivative of a DDE initial function problem}
 {\bf{Input:}} dynamics parameters $\vw$, time delay $\tau$, start time $0$, stop time $T=n\cdot\tau$, final state $\vh(T)$, loss gradient $\partial{L}/\partial{\vh(T)}$\\
 $~~~~$$\dfrac{\partial{L}}{\partial{\vw}} = 0_{|\vw|}$\\
 $~~~~${\bf{for} $i$ in range($n-1, -1, -1$)}:\\
 $~~~~$$~~~~$$s_0=[\vh(T),\vh(T-\tau),..., \vh(\tau),
 \dfrac{\partial{L}}{\partial{\vh(T)}},...,
 \dfrac{\partial{L}}{\partial{\vh((i+1)\cdot{\tau})}}, \dfrac{\partial{L}}{\partial{\vw}}]$\\
 $~~~~$$~~~~${\bf{def}} aug${\underline{~}}$dynamics([$\vh_{n-1}(t)$,...,$\vh_{0}(t)$, $\bm{\lambda}_{n-1}(t)$,..,$\bm{\lambda}_i(t)$, .], $t$, $\vw$):\\
 $~~~~$$~~~~$$~~~~${\bf{return}} [$f_{n-1}(t)$, $f_{n-2}(t)$ $, ..., f_{0}(t),$ 
 $g_{n-1}(t),...,g_{i}(t)$,
 $-\bm{\lambda}_i(t)^{\top} \dfrac{\partial{f_i(t)}}{\partial{\vw}} $]\\
 $~~~~$$~~~~$$
 [\dfrac{\partial{L}}{\partial{\vh(i\cdot{\tau})}}, \dfrac{\partial{L}}{\partial{\vw}}]$=ODESolve($s_0$, aug${\underline{~}}$dynamics, ${\tau}$, $0$, $\vw$)\\
 {\bf{return}} $\dfrac{\partial{L}}{\partial{\vh(0)}}, \dfrac{\partial{L}}{\partial{\vw}}$
\label{alg_01}
\end{algorithm*}
We solve the forward pass of $\vh$ and backward pass for $\vh$, $\bm{\lambda}$ and $\dfrac{{\rm{d}} L}{{\rm{d}} \vw}$ by a piece-wise ODE solver, which is shown in Algorithm \ref{alg_01}.
For simplicity, we denote by $f(t)$ and $g(t)$ the vector filed of $\vh$ and $\bm{\lambda}$, respectively. Moreover, in this paper, we only consider the initial function $\bm{\phi}(t)$ as a constant function, i.e., $\bm{\phi}(t)=\vh_0$. Assume that $T=n\cdot\tau$ and denote $f_k(t) = f(k\cdot\tau + t)$, $g_k(t) = g(k\cdot\tau + t)$ and $\bm{\lambda}_k(t) = \bm{\lambda}(k\cdot\tau+t)$.

In the traditional framework of the NODEs, we can calculate the gradients of the loss function and recompute the hidden states by solving another augmented ODEs in a reversal time duration. However, to achieve the reverse mode of the NDDEs in Algorithm \ref{alg_01}, we need to store the checkpoints of the forward hidden states $h(i\cdot\tau)$ for $i=0,1,...,n$, which, together with the adjoint $\bm{\lambda}(t)$, can help us to recompute $h(t)$ backward in every time periods. The main idea of the Algorithm \ref{alg_01} is to convert the DDEs as a piece-wise ODE such that one can naturally employ the framework of the NODEs to solve it.

{\bf Complexity analysis of the Algorithm \ref{alg_01}}. As shown in \cite{chen2018neural}, the memory and the time costs for solving the NODEs are $\mathcal{O}(1)$ and $\mathcal{O}(L)$, where $L$ is the number of the function evaluations in the time interval $[0, T]$. More precisely, solving the NODEs is memory efficient without storing any intermediate states of the evaluated time points in solving the NODEs. Here, we intend to analyze the complexity of Algorithm \ref{alg_01}. It should be noted that the state-of-the-art for the DDE software is not as advanced as that for the ODE software. Hence, solving a DDE is much more difficult compared with solving the ODE. There exist several DDE solvers, such as the popular solver, the dde23 provided by MATLAB. However, these DDE solvers usually need to store the history states to help the solvers access the past time state $h(t-\tau)$. Hence, the memory cost of DDE solvers is $\mathcal{O}(H)$, where $H$ is the number of the history states. This is the major difference between the DDEs and the ODEs, as solving the ODEs is memory efficient, i.e., $\mathcal{O}(1)$. In Algorithm \ref{alg_01}, we propose a piece-wise ODE solver to solve the DDEs. The underlying idea is to transform the DDEs into the piece-wise ODEs for every $\tau$ time period such that one can naturally employ the framework of the NODEs. More precisely, we compute the state at time $k \tau$ by solving an augmented ODE with the augmented initial state, i.e., concatenating the states at time $-\tau, 0, \tau, 
... , (k-1)\tau$ into a single vector. Such a method has several strengths and weaknesses as well. The strengths include: 
\begin{itemize}
 \item One can easily implement the algorithm using the framework of the NODEs, and
 \item Algorithm \ref{alg_01} becomes quite memory efficient $\mathcal{O}(n)$, where we only need to store a small number of the forward states, $\vh(0),...,\vh(n\tau)$, to help the algorithm compute the adjoint and the gradients in a reverse mode. Here, the final $T$ is assumed to be not very large compared with the time delay $\tau$. For example, in our experiments on image datasets, we set $T=n\tau$ with $n=1$.
\end{itemize}

\subsection{Representation capability of NDDEs}

Here, we use some synthetic datasets produced by typical examples to compare the performance of the NODES and the NDDEs. In \cite{dupont2019augmented}, it is proved that the NODEs cannot represent the function $g:\mathbb{R}^d \rightarrow \mathbb{R}$, defined by
\begin{equation*}
 g(\vx) = \left\{
 \begin{array}{ll}
 1, &\mbox{if}~\|\vx\|\leq r_1,\\
 -1, &\mbox{if}~r_2\leq\|\vx\|\leq r_3,
 \end{array}
 \right.
\end{equation*}
where $0<r_1<r_2<r_3$ and $\|\cdot\|$ is the Euclidean norm. The following proposition shows that the NDDEs have a stronger capability of representation. 

\begin{proposition}
 \label{prop_01}
 The NDDEs can represent the function $g(\vx)$ specified above.
\end{proposition}

To validate this proposition, we construct a special form of the NDDEs by
\begin{equation}
\label{sep_2d_eq}
 \left\{
 \begin{array}{ll}
 \dfrac{{\rm{d}}{\vh_i(t)}}{{\rm{d}} t} = \|\vh_{t-\tau}\| - r, & t>=0 ~ \mbox{and}~i=1, \\
 \dfrac{{\rm{d}}{\vh_i(t)}}{{\rm{d}} t} = 0, & t>=0 ~ \mbox{and}~i=2,...,d, \\
 \vh(t)=\vx, & t<=0.
 \end{array}
 \right.
\end{equation}
where $r := (r_1 + r_2)/2$ is a constant and the final time point $T$ is supposed to be equal to the time delay $\tau$ with some sufficient large value. Under such configurations, we can linearly separate the two clusters by some hyperplane. 

\begin{figure}[htb]
	\centering
	\includegraphics[width=0.48\textwidth]{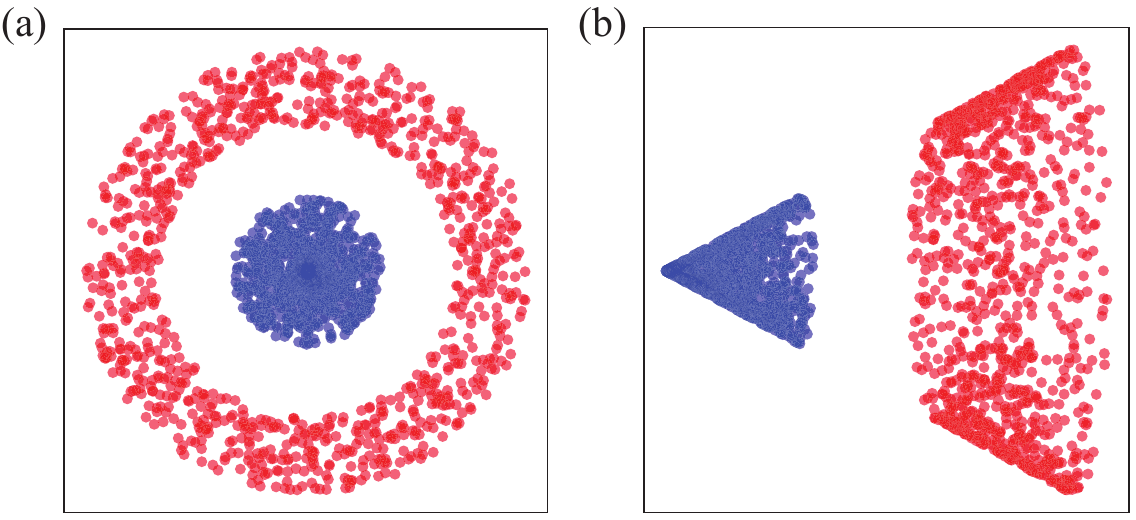}
	\caption{(a) The data at the time $t=0$ and (b) the transformed data at a sufficient large final time $T$ of the DDEs $(\ref{sep_2d_eq})$. Here, the transformed data are linearly separable.}
	\label{sep_fig_2d}
\end{figure}

\begin{figure*}[htb]
 \begin{center}
 \includegraphics[width=18cm]{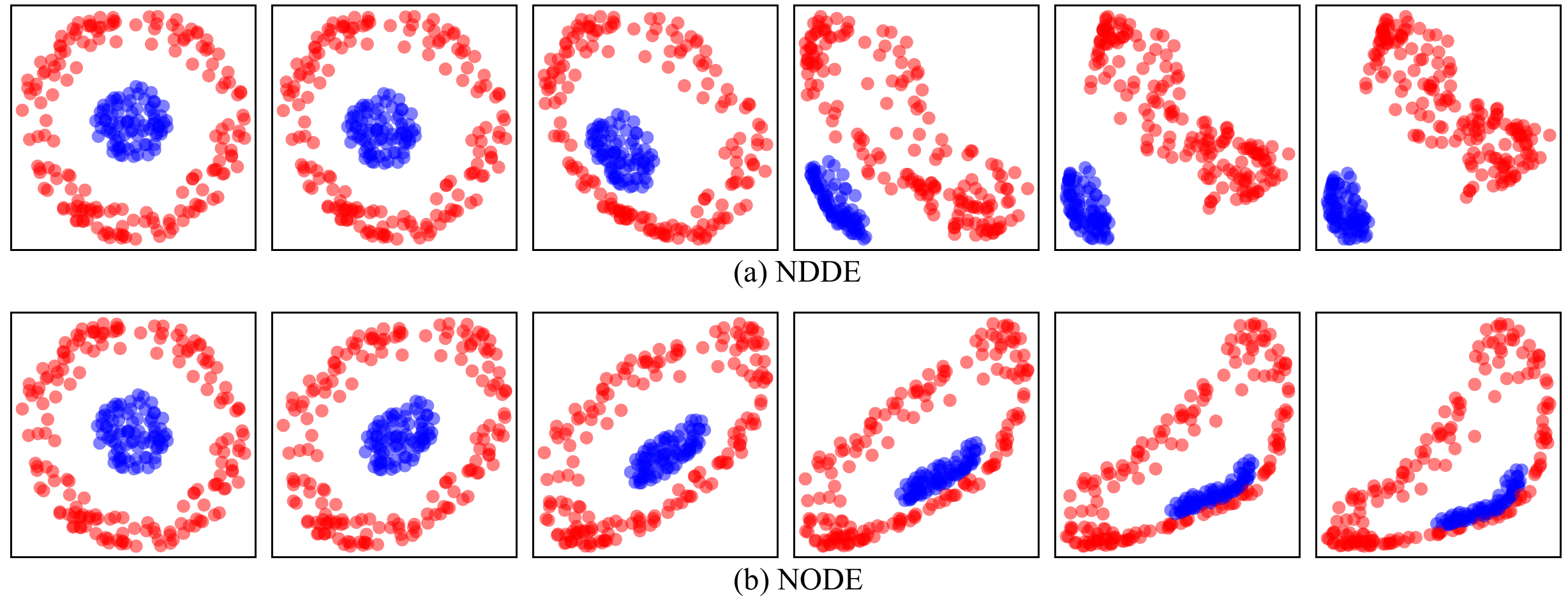}
 \end{center}
 \caption{Evolutions of the NDDEs (a) and the NODEs (b) in the feature space during the training procedure. Here, the evolution of the NODEs is directly produced by the code provided in \cite{dupont2019augmented}.} 
 \label{evolution_train}
\end{figure*}

\begin{figure}[htbp]
 \begin{center}
 \includegraphics[width=8.7cm]{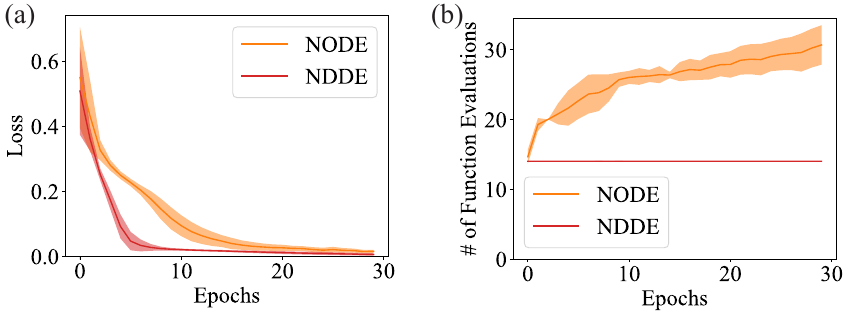}
 \end{center}
 \caption{Presented are the training losses (a) and the number of function evaluations (NFE) (b) of the NODEs and the NDDEs on fitting the function $g(x)$ for $d=2$.} \label{fig_flow_loss}
\end{figure}

\begin{figure*}[htbp]
 \begin{center}
 \includegraphics[width=18cm]{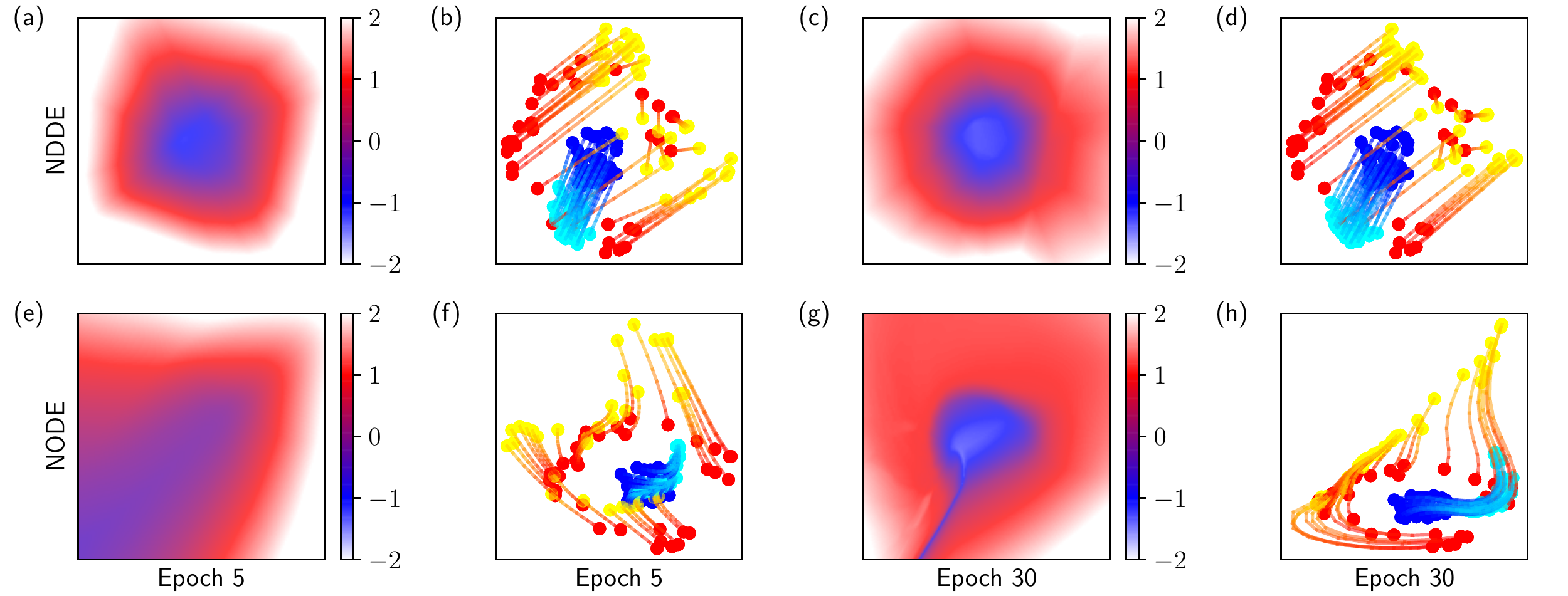}
 \end{center}
 \caption{Decision boundaries and the flows of NODE (a-d) and NDDE (e-h) on a concentric dataset in different training epochs. The flow of the NODEs is generated by the code provided in \cite{dupont2019augmented}.} \label{fig_flow}
\end{figure*}

 In Fig. $\ref{sep_fig_2d}$, we, respectively, show the original data of $g(\vx)$ for the dimension $d=2$ and the transformed data by the DDEs $(\ref{sep_2d_eq})$ with the parameters $r_1=1$, $r_2=2$, $r_3=3$, $T=10$, and $\tau=10$. Clearly, the transformed data by the DDEs are linearly separable.

We also train the NDDEs to represent the $g(\vx)$ for $d=2$, whose evolutions during the training procedure are, respectively, shown in Fig. $\ref{evolution_train}$. This figure also includes the evolution of the NODEs. 
Notably, while the NODEs is struggled to break apart the annulus, the NDDEs easily separate them. The training losses and the flows of the NODEs and the NDDEs are depicted, respectively, in Fig.~$\ref{fig_flow}$. Particularly, the NDDEs achieve lower losses with the faster speed and directly separate the two clusters in the original $2$-D space; however, the NODEs achieve it only by increasing the dimension of the data and separating them in a higher-dimensional space.

 In general, we have the following theoretical result for the NDDEs, whose proof is provided in Appendix. 
\begin{theorem} 
 \label{thm_02}
 (Universal approximating capability of the NDDEs). For any given continuous function $F: \mathbb{R}^n \rightarrow \mathbb{R}^n$, if one can construct a neural network for approximating the map $G(\vx) = \dfrac{1}{T}[F(\vx) - \vx]$, then there exists an NDDE of $n$-dimension that can model the map $\vx \mapsto F(\vx)$, that is, $\vh(T)\approx F(\vx)$ with the initial function $\bm{\phi}(t)=\vx$ for $t\leq0$. 
\end{theorem}

\section{Illustrative Experiments}
\subsection{System reconstruction}
\begin{figure*}[htb]
 \begin{center}
 \includegraphics[width=18cm]{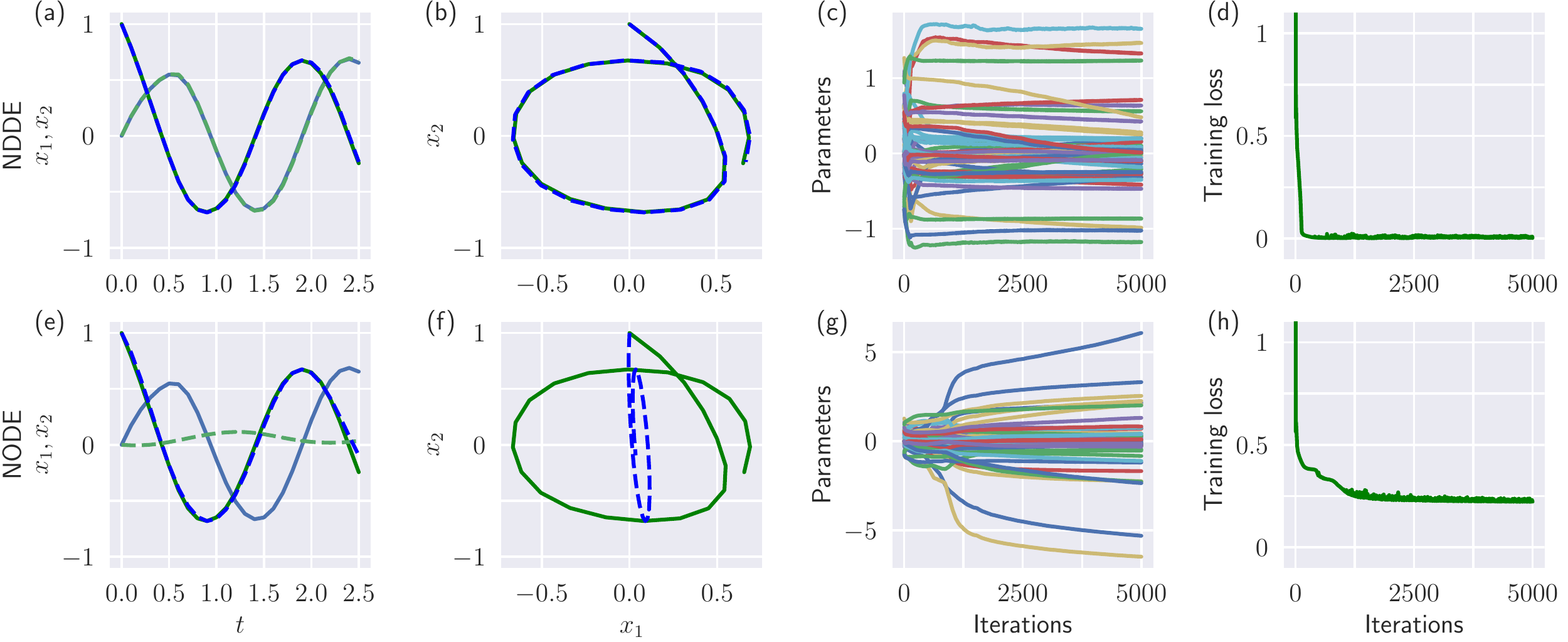}
 \end{center}
 \caption{Comparison of the NDDEs (a-d) versus the NODEs (e-h) in fitting a 2-D time series with the delay effect in the original system. From the left to the right, the true and the fitted time series (a, e), the true and the fitted trajectories in the phase spaces (b, f), the dynamics of the parameters in the neural networks (c, g), and the dynamics of the losses during the training processes (d, f).
 }
 \label{spiral_2d_fig}
\end{figure*}

{\bf System reconstruction with known delays in a model-free case.}
Additionally, NDDEs are suitable for fitting the time series with the delay effect in the original systems, which cannot be easily achieved by using the NODEs. To illustrate this, we use a model of $2$-D DDEs, written as $\dot{\vx} = \mA \tanh(\vx(t) + \vx(t - \tau))$ with $\vx(t)={\vx}_0$ for $t<0$. Given the time series generated by the system, we use the NDDEs and the NODEs to fit it, respectively. Figure $\ref{spiral_2d_fig}$ shows that the NNDEs approach a much lower loss, compared to the NODEs. More interestingly, the NODEs prefer to fit the dimension of ${\vx}_2$ and its loss always sustains at a larger value, e.g. $0.25$ in Figure $\ref{spiral_2d_fig}$. The main reason is that two different trajectories generated by the autonomous ODEs cannot intersect with each other in the phase space, due to the uniqueness of the solution of ODEs.

\begin{figure*}[htb]
 \begin{center}
 \includegraphics[width=18cm]{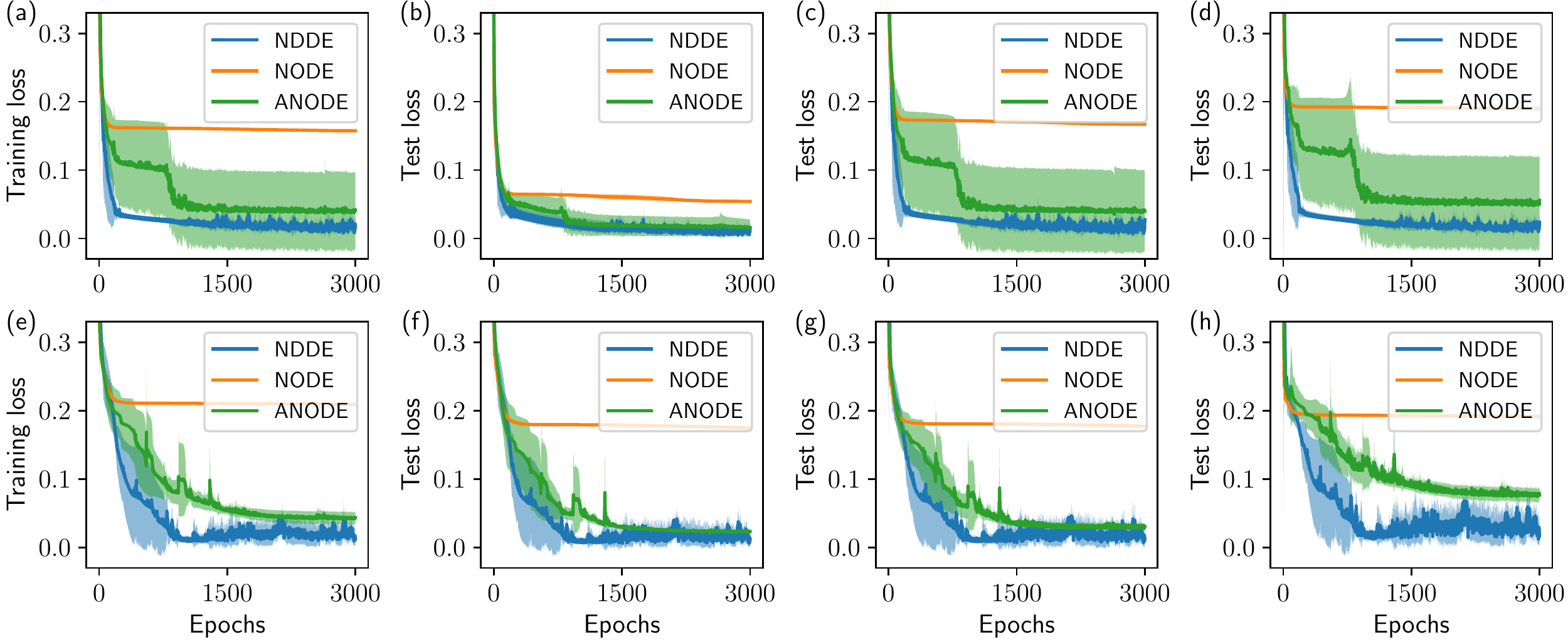}
 \end{center}
 \caption{
 {The training losses and the test losses of the population dynamics (a-d) and the Mackey-Glass system (e-h) by using the NDDEs, the NODEs and the ANODEs (where the augmented dimension equals to $1$). The first column (a, e) in the figures is the training losses. The figures from the second column to the fourth column present the test losses over the time intervals of the length $\tau$ (b, f), $2\tau$ (c, g), $5\tau$ (d, h). The delays for the two DDEs are both designed as 1, respectively. The other parameters are set as $r=1.8$, $\beta=4.0$, $n=9.65$, and $\gamma=2$.}
 }
 \label{P_M_fig}
\end{figure*}

We perform experiments on another two classical DDEs, i.e., the population dynamics and the Mackey-Glass system \cite{erneux2009applied}. Specifically, the equation of the dimensionless population dynamics is $\dot{x} = r x(t) (1- x(t-\tau))$, where $x(t)$ is the ratio of the population to the carrying capacity of the population, $r$ is the growth rate and $\tau$ is the time delay. The Mackey-Glass system is written as $\dot{x} = \beta \dfrac{ x(t-\tau)}{1 + x^n(t-\tau)} - \gamma x(t)$, where $x(t)$ is the number of the blood cells, $\beta, n, \gamma$, and $\tau$ are the parameters of biological significance. The NODEs and the NDDEs are tested on these two dynamics. As shown in Figure \ref{P_M_fig}, a very low training loss is achieved for the NDDEs while the loss of the NODEs does not go down afterward, always sustaining at some larger value. As for the predicting durations, the NDDEs thereby achieve a better performance than the NODEs. The details of the training could be found in Appendix.

\begin{figure*}[htb]
 \begin{center}
 \includegraphics[width=18cm]{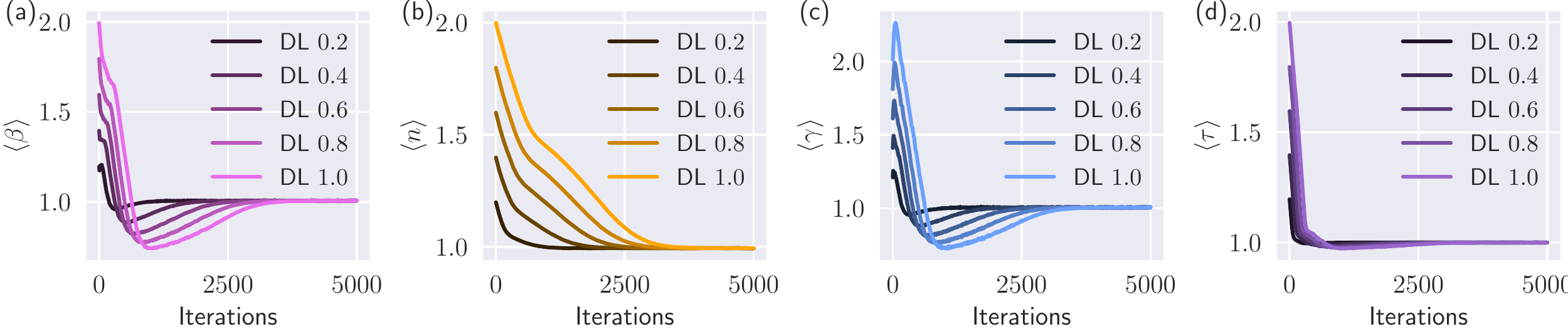} 
 \end{center}
 \caption{System identification of the Mackey-Glass system by using the NDDEs. The identified $\beta$, $n$, $\gamma$, and $\tau$ are shown in panels (a), (b), (c), and (d), respectively. Here, we normalize the learned parameter at different training epochs by dividing the true value. Specifically, these parameters are initialized in different deviation levels, i.e., $p=(1 + DL)p$, where $p\equiv \beta, n, \gamma$, or $\tau$, and $DL$ is selected from the set $\{0.1, 0.2, ..., 1.0\}$. 
 }
 \label{MG_Identification_fig}
\end{figure*}

\begin{figure}[htb]
 \begin{center}
 \includegraphics[width=9cm]{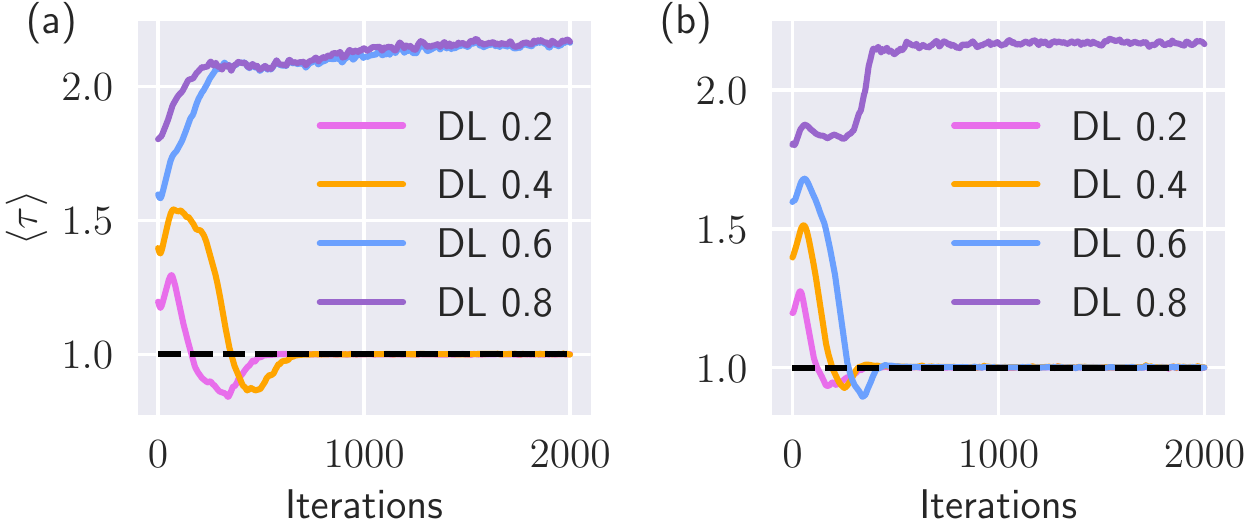}
 \end{center}
 \caption{Infering the underlying time delay of the Mackey-Glass system by using the NDDEs in a model-free manner.  The learned time delays are shown in (a) and (b) with $8$ and $16$ hidden neurons, respectively,  and we use $3$ hidden layers and the $\tanh$ activation function. $DL$ is selected from the set $\{0.2, 0.4,0.6, 0.8\}$. 
 }
 \label{MG_Identification_model_free_fig}
\end{figure}

{\bf System reconstruction with unknown delays in a model-free or model-based case.} For the Mackey-Glass system, we employ the NDDEs framework in a model-based manner to identify the unknown biological parameters, including $\beta, n, \gamma$, and $\tau$, solely from the observable time series. As shown in Fig. \ref{MG_Identification_fig}, under quite a range of the initialized parameters, these parameters can be convergent to the true values. In addition, one can directly infer the underlying time delay by using the NDDEs in a model-free manner as shown in  Fig. \ref{MG_Identification_model_free_fig} while it may fail in a large initialized time delay. In this example, we set  $\beta=2, n=10, \gamma=1$, and $\tau=3.18$, which is used in Refs. \cite{zhu2019detecting,zhu2023leveraging}.


\subsection{Experiments on image datasets}

For the image datasets, we not only use the NODEs and the NDDEs but also an extension of the NDDEs, called the NODE+NDDE, which treats the initial function as an ODE. In our experiments, such a model exhibits the best performance, revealing strong capabilities in modeling and feature representations. Moreover, inspired by the idea of the augmented NODEs \cite{dupont2019augmented}, we extend the NDDEs and the NODE+NDDE to the A+NDDE and the A+NODE+NDDE, respectively. Precisely, for the augmented models, we augment the original image space to a higher-dimensional space, i.e., $\mathbb{R}^{c\times h \times w} \rightarrow \mathbb{R}^{ (c+p)\times h \times w}$, where $c$, $h$, $w$, and $p$ are, respectively, the number of channels, height, width of the image, and the augmented dimension. With such configurations of the same augmented dimension and approximately the same number of the model parameters, comparison studies on the image datasets using different models are reasonable.
For the NODEs, we model the vector filed $f(\vh(t))$ as the convolutional architectures together with a slightly different hyperparameter setups in \cite{dupont2019augmented}. The initial hidden state is set as $\vh(0)\in \mathbb{R}^{c\times h \times w}$ with respect to an image. For the NDDEs, we design the vector filed as $f({\rm concat}(\vh(t), \vh(t-\tau)))$, mapping the space from $\mathbb{R}^{2c\times h \times w}$ to $\mathbb{R}^{c\times h \times w}$, where ${\rm concat}(\cdot, \cdot)$ executes the concatenation of two tensors on the channel dimension. Its initial function is designed as a constant function, i.e., $h(t)$ is identical to the input/image for $t<0$. For the NODE+NDDE, we model the initial function as an ODE, which follows the same model structure of the NODEs. For the augmented models, the augmented dimension is chosen from the set \{1, 2, 4\}. Moreover, the training details could be found in Appendix, including the training setting and the number of function evaluations for each model on the image datasets.

\begin{figure*}[htb]
 \begin{center}
 \includegraphics[width=18cm]{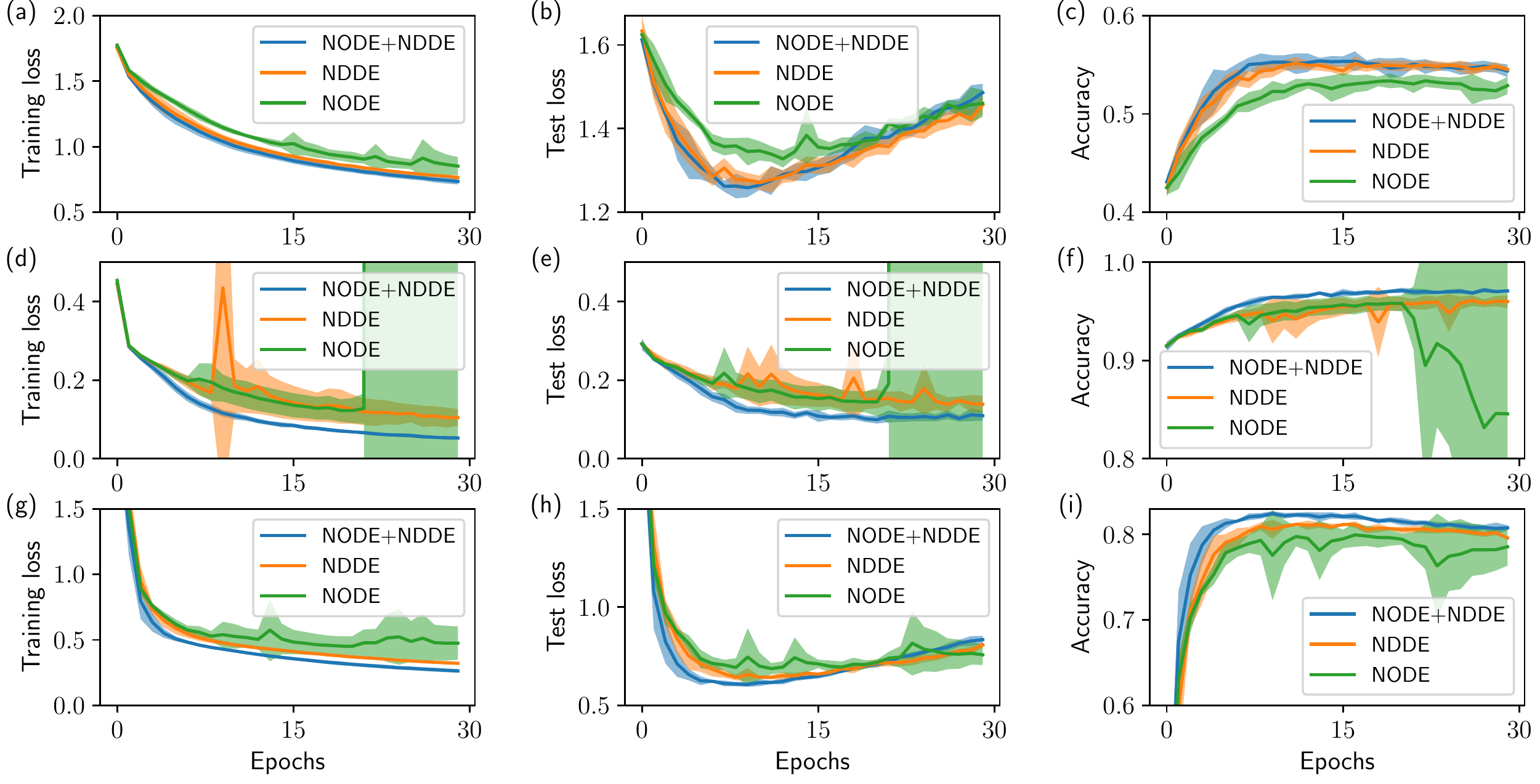}
 \end{center}
 \caption{The training loss (left column), the test loss (middle column), and the accuracy (right column) over 5 realizations for the three image sets, i.e., CIFAR10 (top row), MNIST (middle row), and SVHN (bottom row). For each model, the initial (resp. final) time is set as 0 (resp. 1), and the delays of the NDDEs and its extensions are all set as 1, simply equal to the final time.
 }
 \label{mnist_cifar_fig}
\end{figure*}

The training processes on MNIST, CIFAR10, and SVHN are shown in Fig.~\ref{mnist_cifar_fig}. Overall, the NDDEs and its extensions have their training losses decreasing faster than the NODEs/ANODEs which achieve lower training and test loss. Also, the test accuracies are much higher than that of the NODEs/ANODEs (refer to Tab.~\ref{table}). Naturally, the better performance of the NDDEs is attributed to the integration of the information not only on the hidden states at the current time $t$ but at the previous time $t-\tau$. This kind of framework is akin to the key idea proposed in \cite{huang2017densely}, where the information is processed on many hidden states. Here, we run all the experiments for 5 times independently. 

\begin{figure}[htb]
\centering
 \includegraphics[width=8.9cm]{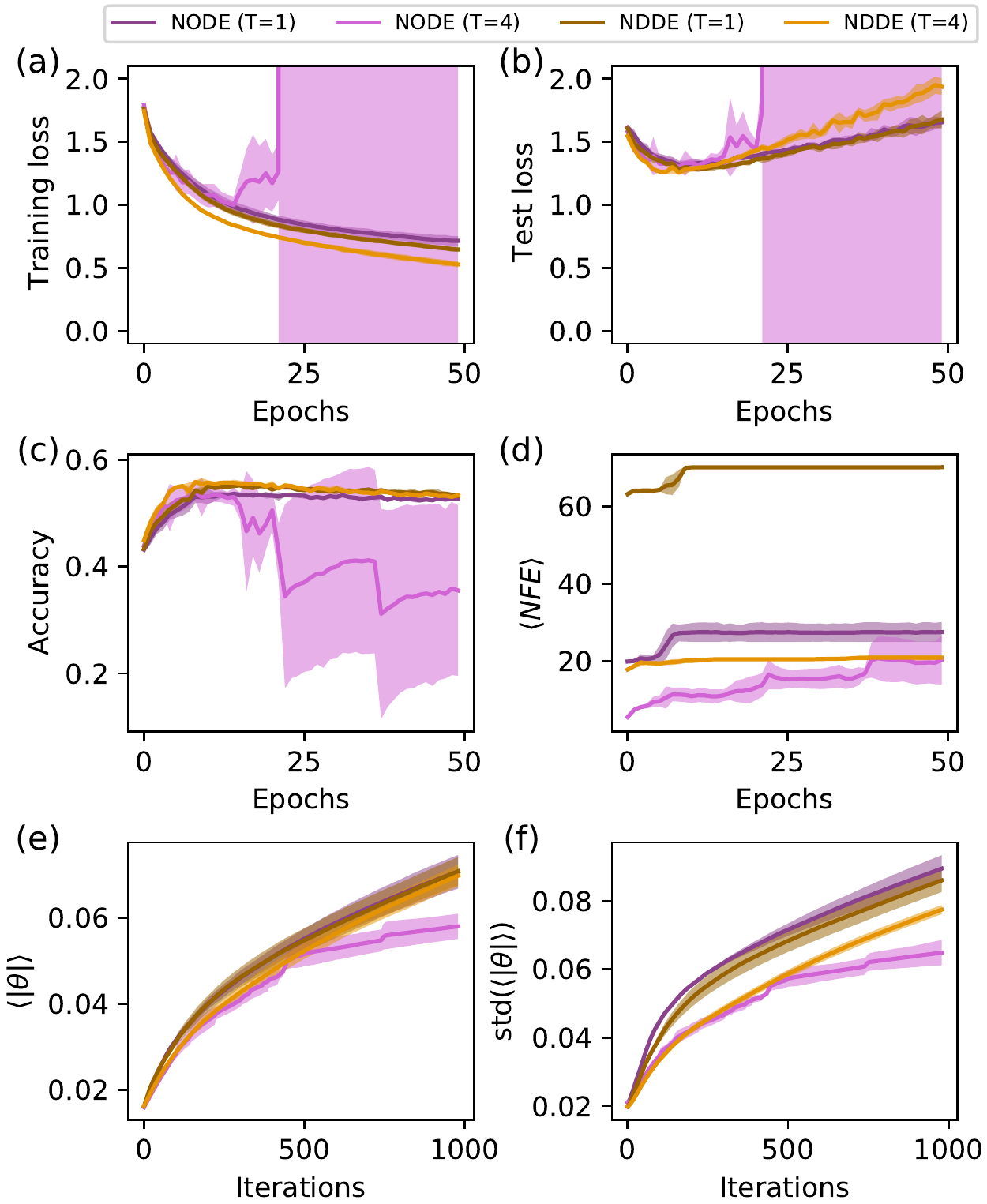}

 \caption{The training loss (a), the test loss (b), the accuracy (c), the normalized NFE (i.e., $\langle NFE \rangle = NFE / T$) (d), the mean value (e) and the standard deviation (f) of $|\vw|$ over 4 realizations for NODEs and NDDEs with different terminal time $T$ ($1$ or $4$) on CIFAR10. For the NDDE, we choose $\tau \equiv T$. 
 }
 \label{cifar_diff_T_fig}
\end{figure}

In addition, we conduct the experiment by using the NODEs and NDDEs with different fixed or learnable terminal time $T$ (1 or 4) on Cifar10. As shown in Tab. \ref{table_adap} and Fig. \ref{cifar_diff_T_fig}, a large  fixed or learnable terminal time $T$ can facilitate the training and the test performance, including the fast training, the high accuracy, and the low normalized NFE. Intuitively, from the perspective of the control field, when $T$ becomes large, the learnable parameters, served as a controller, do not require a significant change to experimentally control the task performance., which can be observed from the Fig. \ref{cifar_diff_T_fig}(f) with a low standard deviation of $|\vw|$.

{
\begin{table*}[h!]
\centering
\captionsetup[subtable]{position = below}
\captionsetup[table]{position=top}
 \begin{tabular}{llll}
 \hline\hline
 \multicolumn{1}{c}{\bf ~} &\multicolumn{1}{c}{CIFAR10} &\multicolumn{1}{c}{MNIST} &\multicolumn{1}{c}{SVHN}
 \\ \hline 
NODE &$53.92\%\pm 0.67$ &$96.21\%\pm 0.66$ &$80.66\%\pm 0.56$\\
NDDE &$55.69\%\pm 0.39$ &$96.22\%\pm 0.55$ &$81.49\%\pm 0.09$\\
NODE+NDDE &${\bf 55.89\%\pm 0.71}$ &${\bf 97.26\%\pm 0.22}$ &${\bf 82.60\%\pm 0.22}$\\
\hline
A1+NODE &$56.14\%\pm 0.48$ &$97.89\%\pm 0.14$ &$81.17\%\pm 0.29$\\
A1+NDDE &$56.83\%\pm 0.60$ &$97.83\%\pm 0.07$ &$82.46\%\pm 0.28$\\
A1+NODE+NDDE &${\bf 57.31\%\pm 0.61}$ &${\bf 98.16\%\pm 0.07}$ &${\bf 83.02\%\pm 0.37}$\\
\hline
A2+NODE &$57.27\%\pm 0.46$ &$98.25\%\pm 0.08$ &$81.73\%\pm 0.92$\\
A2+NDDE &$58.13\%\pm 0.32$ &$98.22\%\pm 0.04$ &$82.43\%\pm 0.26$\\
A2+NODE+NDDE &${\bf 58.40\%\pm 0.31}$ &${\bf 98.26\%\pm 0.06}$ &${\bf 83.73\%\pm 0.72}$\\
\hline
A4+NODE &$58.93\%\pm 0.33$ &$98.33\%\pm 0.12$ &$82.72\%\pm 0.60$\\
A4+NDDE &$59.35\%\pm 0.48$ &$98.31\%\pm 0.03$ &$82.87\%\pm 0.55$\\
A4+NODE+NDDE &${\bf 59.94\%\pm 0.66}$ &${\bf 98.52\%\pm 0.11}$ &${\bf 83.62\%\pm 0.51}$\\
 \hline\hline
 \end{tabular}
 \caption{The test accuracies with their standard deviations over 5 realizations on the three image datasets. In the first column, $p$ (=1, 2, or 4) in A$p$ means the number of the channels of zeros into the input image during the augmentation of the image space $\mathbb{R}^{c\times h \times w} \rightarrow \mathbb{R}^{ (c+p)\times h \times w}$ \cite{dupont2019augmented}. For each model, the initial (resp. final) time is set as 0 (resp. 1), and the delays of the NDDEs and its extensions are all set as 1, simply equal to the final time.
 }
 \label{table}
\end{table*}
}

\begin{table}[htb]
 \begin{center}
\resizebox{\linewidth}{!}{
 \begin{tabular}{llll}
 \hline\hline
 \multicolumn{1}{c}{\bf ~} &\multicolumn{1}{c}{CIFAR10} &\multicolumn{1}{c}{MNIST} &\multicolumn{1}{c}{SVHN}
 \\ \hline 
NODE &$92/107645$ &$92/84395$ &$92/107645$\\
NDDE &$92/107921$ &$92/84487$ &$92/107921$\\
NODE+NDDE &$64/105680$ &$64/82156$ &$64/105680$\\
\hline
A1+NODE &$86/108398$ &$87/84335$ &$86/108398$\\
A1+NDDE &$85/107189$ &$86/82944$ &$85/107189$\\
A1+NODE+NDDE &$60/107218$ &$61/83526$ &$60/107218$\\
\hline
A2+NODE &$79/108332$ &$81/83230$ &$79/108332$\\
A2+NDDE &$78/107297$ &$81/83473$ &$78/107297$\\
A2+NODE+NDDE &$55/107265$ &$57/83101$ &$55/107265$\\
\hline
A4+NODE &$63/108426$ &$70/84155$ &$63/108426$\\
A4+NDDE &$62/107719$ &$69/83237$ &$62/107719$\\
A4+NODE+NDDE &$43/106663$ &$49/83859$ &$43/106663$\\
 \hline\hline
 \end{tabular}
}
 \end{center}
 \caption{The number of the filters and the whole parameters in each model used for CIFAR10, MNIST, and SVHN. In the first column, A$p$ with $p=1,2,4$ indicates the augmentation of the image space $\mathbb{R}^{c\times h \times w} \rightarrow \mathbb{R}^{ (c+p)\times h \times w}$.
 }
 \label{table2}
\end{table}

\begin{table*}[htb]
\centering
\captionsetup[subtable]{position = below}
\captionsetup[table]{position=top}
 \begin{tabular}{l|ll|ll}
 \hline\hline
\multicolumn{1}{c|}{\bf ~} &\multicolumn{1}{c}{T1+Adap0} &\multicolumn{1}{c|}{T1+Adap1} &\multicolumn{1}{c}{T4+Adap0} &\multicolumn{1}{c}{T4+Adap1}\\ \hline
NODE & $54.12\%\pm 0.35$& $53.95\%\pm 1.00$& $54.52\%\pm 1.46$& $54.12\%\pm 0.33$\\
NDDE & ${\bf 55.57\%\pm 0.15}$& ${\bf 55.72\%\pm 0.23}$& ${\bf 56.41\%\pm 0.20}$& ${\bf 56.15\%\pm 0.18}$\\
\hline\hline
\end{tabular}
\caption{The test accuracies with their standard deviations over 4 realizations on Cifar10. In the table, $i$ ($1$ or $4$) in T$i$ means the terminal time $T=i$ while $j=0$ (resp., $1$) in Adap$j$ means that $T$ is fixed (resp., learnable) parameter.
 }
 \label{table_adap}
\end{table*}

\section{Discussion}

In this section, we present the limitations of the NDDEs and further suggest several potential directions for future works. We add the delay effect to the NDDEs, which renders the model absolutely irreversible. Algorithm \ref{alg_01} thus requires a storage of the checkpoints of the hidden state $\vh(t)$ at every instant of the multiple of $\tau$. Actually, solving the DDEs is transformed into solving the ODEs of an increasingly high dimension with respect to the ratio of the final time $T$ and the time delay $\tau$. This definitely indicates a high computational cost. To further apply and improve the framework of the NDDEs, a few potential directions for future works are suggested, including:

{\bf Applications to more real-world datasets.} In the real-world systems such as physical, chemical, biological, and ecological systems, the delay effects are inevitably omnipresent, truly affecting the dynamics of the produced time-series \cite{bocharov2000numerical, kajiwara2012construction}. The NDDEs are undoubtedly suitable for realizing model-free and accurate prediction \cite{quaglino2019snode}. 
Additionally, since the NODEs have been generalized to the areas of Computer Vision and Natural Language Processing \cite{he2019ode, yang2019pointflow, liu2020learning}, the framework of the NDDEs probably can be applied to the analogous areas, where the delay effects should be ubiquitous in those streaming data.

{\bf Extension of the NDDEs.} A single constant time delay in the NDDEs can be further generalized to the case of multiple or/and distributed time delays \cite{shampine2001solving}. As such, the model is likely to have a much stronger capability to extract the feature, because the model leverages the information at different time points to make the decision in time. 
All these extensions could be potentially suitable for some complex tasks. However, such a complex model may require a tremendously huge computational cost.

{\bf Time-dependent controllers.} From a viewpoint of control theory, the parameters in the NODEs/NDDEs could be regarded as time-independent controllers, viz. constant controllers. A natural generalization way is to model the parameters as time-dependent controllers. In fact, such controllers were proposed in \cite{zhang2019anodev2}, where the parameters $\vw(t)$ were treated as another learnable ODE $\dot{\vw} = q(\vw(t), \vp)$, $q(\cdot, \cdot)$ is a different neural network, and the parameters $\vp$ and the initial state $\vw(0)$ are pending for optimization. Also, the idea of using a neural network to generate the other one was initially conceived in some earlier works including the study of the hypernetworks \cite{ha2016hypernetworks}.

\section{Conclusion}

In this article, we establish the framework of NDDEs, whose vector fields are determined mainly by the states at the previous time. We employ the adjoint sensitivity method to compute the gradients of the loss function. The obtained adjoint dynamics backward follow another DDEs coupled with the forward hidden states. We show that the NDDEs can represent some typical functions that cannot be represented by the original framework of NODEs. Moreover, we have validated analytically that the NDDEs possess the universal approximating capability. We also demonstrate the exceptional efficacy of the proposed framework by using the synthetic data or real-world datasets. All these reveal that integrating the elements of dynamical systems into the architecture of neural networks could be potentially beneficial to the promotion of network performance.

\subsubsection*{Acknowledgments}
Q.Z. was supported by the China Postdoctoral Science Foundation (No. 2022M720817), by the Shanghai Postdoctoral Excellence
Program (No. 2021091), and by the STCSM (Nos. 21511100200 and 22ZR1407300). W.L. was supported by the National Natural Science
Foundation of China (No. 11925103) and by the STCSM (Nos. 22JC1402500, 22JC1401402, and 2021SHZDZX0103).

\bibliographystyle{IEEEtran}
\bibliography{main}
\end{document}